\documentclass[11pt,a4paper]{article}
\usepackage[T1]{fontenc}
\usepackage[utf8]{inputenc}
\usepackage{mathptmx}
\usepackage[scaled=0.9]{helvet}
\usepackage{courier}
\usepackage[margin=2.5cm]{geometry}
\usepackage{amsmath,amssymb}
\usepackage{graphicx}
\usepackage{booktabs,tabularx,array}
\usepackage[numbers,sort&compress]{natbib}
\usepackage{caption}
\usepackage{enumitem}
\usepackage{microtype}
\usepackage{xcolor}
\usepackage[colorlinks=true,linkcolor=blue!50!black,citecolor=blue!50!black,urlcolor=blue!50!black]{hyperref}

\newcolumntype{Y}{>{\raggedright\arraybackslash}X}
\setlist{itemsep=2pt,topsep=4pt}

\title{Does Validation Loss Predict Real-Robot Success?\\ Evidence from Four Imitation-Learning Policies\\ on a Low-Cost Arm}
\author{Simon Cowling\\[2pt] fysical.ai, Denmark\\[2pt] \texttt{simon@fysical.ai}}
\date{30 October 2026}

\hypersetup{pdftitle={Does Validation Loss Predict Real-Robot Success? Evidence from Four Imitation-Learning Policies on a Low-Cost Arm},pdfauthor={Simon Cowling}}

\begin{document}
\maketitle

\begin{abstract}
Held-out validation loss is the standard offline signal for choosing checkpoints and comparing models, but its relation to closed-loop success on a real robot is rarely measured. We measure it for ACT, Diffusion Policy, SmolVLA and $\pi_{0.5}$ (action expert only), trained on nested subsets of 12 to 200 demonstrations of a pick-and-place task on an SO-101 arm. We score 82 checkpoints on 25 held-out demonstrations, with each policy's training loss and with the error of the actions it outputs, and compare the scores with 930 bench rollouts. We also score all 700 intermediate checkpoints of 70 training runs, and recover the ball positions of every training, held-out and bench episode from the front camera. Within each policy, hold-out action error correlates with bench success (Spearman $\rho$ from $-0.23$ to $-0.88$), but no more strongly than the number of training episodes alone, and not at all among seeds trained on the same data. Across policies the ranking is inverted ($\rho$ = +0.90 at 200 episodes): the policy with the lowest hold-out error succeeded in 3 of 110 rollouts. The most common failure, the arm never leaving its rest pose, ended 96\% of rollouts for Diffusion Policy trained on untrimmed demonstrations and 47\% for SmolVLA, and cannot occur in a hold-out set made of demonstrations. Among rollouts in which the arm moved, SmolVLA's hold-out action error predicts success well ($\rho$ = $-0.81$). For the diffusion and flow-matching policies, hold-out loss rises two- to five-fold during training while hold-out action error falls by up to 29\%. Held-out metrics are a sound diagnostic of learning but no substitute for evaluation on the robot, and we recommend reporting initiation failure and success among started rollouts as separate bench outcomes.
\end{abstract}

\section{Introduction}\label{sec:1}

Every rollout on a real robot costs minutes of attention and a reset by hand, so a number that could stand in for the bench would save most of the cost of a study. Validation loss is the obvious candidate. It is cheap, it is computed by the same code that trains the model, and in supervised learning it is the standard way to choose a checkpoint and to compare models.

Robot imitation learning breaks two of the assumptions that make it work elsewhere. The policy is trained on single steps but judged on a closed-loop run in which its own actions shape what it sees next, so small errors compound~\cite{ross2011dagger}. And the demonstrations it is scored on always show the task already under way, so some ways of failing never appear in them. Earlier work has found that validation loss is a weak guide to success in this setting~\cite{mandlekar2021robomimic,barannikov2024validation,huang2026cimse}; this study asks how weak, why, and what can replace it.

We answer these questions with the models and bench results of a previous scaling study~\cite{cowling2026scaling}, in which ACT~\cite{zhao2023act}, Diffusion Policy~\cite{chi2023diffusion}, SmolVLA~\cite{shukor2025smolvla} and $\pi_{0.5}$~\cite{black2025pi05} were trained on 12 to 200 demonstrations of one task on one low-cost arm. The contributions are:

\begin{enumerate}
\item A held-out scoring method that works the same way for four very different policy families, and a common measure, action error, that puts them on one scale.
\item A comparison of validation numbers with 930 bench rollouts, within models, across models, across seeds and across two interventions (EMA weights and trimming the idle lead-in of demonstrations).
\item Training curves for 70 runs that settle whether the models were trained long enough, and show that for diffusion and flow-matching policies the loss and the action error move in opposite directions.
\item A map of where the balls were placed in every training, hold-out and bench episode, recovered from the front camera, and a first look at whether a placement's distance from the training data predicts its success.
\end{enumerate}

\section{Background}\label{sec:2}

\subsection{Validation loss}\label{sec:2.1}

A policy learns to answer one question: given what the cameras and joint sensors show now, what did the demonstrator do next? The training loss measures how wrong those answers are on the frames used for training, which a large network can get right by memorising. Validation loss asks the same question on demonstrations kept out of training. In this paper the 25 held-out demonstrations are called the \textbf{hold-out}, and the same calculation on 12 demonstrations every model trained on is called the \textbf{training probe}.

\subsection{What each policy's loss measures}\label{sec:2.2}

The four policies are trained on different objectives, so their losses are not on a common scale and cannot be compared between models.

\begin{table}[tbp]
\centering
\small
\caption{Training objective of each policy and what its loss compares.}
\label{tab:losses}
\begin{tabularx}{\linewidth}{@{}>{\raggedright\arraybackslash}p{2.3cm}YYY@{}}
\toprule
\textbf{Policy} & \textbf{Training objective} & \textbf{What the loss compares} & \textbf{Randomness at scoring time} \\
\midrule
ACT & L1 regression of an action chunk (plus a KL term in training only) & predicted actions with demonstrated actions, normalised & none: deterministic in evaluation mode \\
Diffusion Policy & denoising (DDPM noise prediction, MSE) & the noise the network predicts with the noise that was added to a demonstrated chunk & noise level and noise sample; averaged over 4 fixed-seed draws \\
SmolVLA & flow matching (MSE on velocity) & the predicted direction back to the clean chunk with the true direction & time step and noise sample; 4 draws \\
$\pi_{0.5}$ (action expert only) & flow matching (MSE on velocity) & as SmolVLA & as SmolVLA \\
\bottomrule
\end{tabularx}
\end{table}

Only ACT's loss is itself an error in the actions. For the other three, which are trained by denoising diffusion~\cite{ho2020ddpm} or flow matching~\cite{lipman2023flow}, the loss scores how well the network repairs a deliberately corrupted demonstration, averaged over every level of corruption. That is not the same as the quality of the action the policy finally outputs.

\subsection{Action error: one measure for every model}\label{sec:2.3}

To compare models we also score the action each policy actually produces. The policy is run as it is on the robot (its inference path and saved postprocessor), and its predicted chunk is compared with the demonstrated actions over the steps the robot would execute before re-planning:

\begin{equation}
\mathrm{MAE} = \frac{1}{N\,H\,J}\sum_{n=1}^{N}\sum_{h=1}^{H}\sum_{j=1}^{J}\left|\hat a_{n,h,j}-a_{n,h,j}\right|
\label{eq:mae}
\end{equation}

Here N is the number of scored frames, H the execution horizon, J the six joints, and errors are in the recorded joint units. Horizons differ by policy (ACT 100 steps, SmolVLA and $\pi_{0.5}$ 50, Diffusion Policy 24), and errors grow with distance into the future, so action error is comparable within a policy and between SmolVLA and $\pi_{0.5}$, but levels across other pairs carry a horizon confound (Section~\ref{sec:9}).

\subsection{Related work}\label{sec:2.4}

\textbf{Offline evaluation of robot policies.} The robomimic study found that the checkpoint with the lowest validation loss could perform 50 to 100\% worse on the robot than the best checkpoint~\cite{mandlekar2021robomimic}. A LeRobot community experiment found Diffusion Policy's validation loss rising while its simulated success improved, and ACT's staying flat while its success rose by more than 15\%~\cite{barannikov2024validation}. Restricting offline action error to the task-critical segments of a demonstration improves its correlation with rollout success~\cite{huang2026cimse}. Other work replaces held-out loss with a different offline signal: off-policy classification predicts the relative real-world performance of grasping policies trained in simulation~\cite{irpan2019opc}, and matched simulated environments have been used as a proxy for real-world evaluation of generalist manipulation policies~\cite{li2024simpler}. The present study measures the gap between held-out metrics and real-robot success for four current policy families on one task, and decomposes it by failure mode.

\textbf{Why held-out error and closed-loop success diverge.} A policy trained by behaviour cloning is scored on states drawn from the demonstrations but deployed on states induced by its own actions, where errors compound~\cite{ross2011dagger}. It can also rely on features that are correlated with the expert's action without causing it~\cite{dehaan2019causal}. Policies given a history of observations can learn to repeat the expert's previous action instead of predicting the next one~\cite{wen2020copycat}, and driving policies trained on data in which a stopped vehicle usually stays stopped can fail to start moving again, the inertia problem~\cite{codevilla2019limitations}. The initiation failure analysed in Section~\ref{sec:5.4} has the same form. For diffusion policies trained on small datasets, the network behaves much like a lookup table of training actions~\cite{he2025demystifying}, which Section~\ref{sec:6.1} uses to explain why their hold-out loss and action error diverge.

\textbf{Data scaling in imitation learning.} Lin et al. found that the generalisation of imitation policies to new environments and objects follows a rough power law in their number, and depends far more on that diversity than on the absolute number of demonstrations~\cite{lin2025scaling}. The placement maps in Section~\ref{sec:7} are a small-scale analogue: what grows with the budget is how densely the training placements cover the workspace.

\section{Experimental setup}\label{sec:3}

The models, demonstrations and bench protocol are those of~\cite{cowling2026scaling}, which reports the full training recipes and bench results. This section summarises what the validation analysis needs.

\subsection{Task and robot}\label{sec:3.1}

An SO-101 arm~\cite{so101} picks a green ball, placed among two red distractor balls on a 50 mm grid board, and drops it in a basket. Each episode is recorded at 30 fps from a fixed low front camera and a wrist camera (640 $\times$ 480), with the six joint positions and commanded actions.

\subsection{Demonstrations, hold-out and budgets}\label{sec:3.2}

The pool holds 225 teleoperated demonstrations, recorded in nine batches of 25 under slightly different lighting. A frozen hold-out of 25 episodes was drawn stratified across the nine batches (data seed 20260922) and never used for training. From the other 200, nested training subsets were drawn once and reused for every model and seed: S12 $\subset$ S25 $\subset$ S50 $\subset$ S100 $\subset$ S200. Because S12 sits inside every subset, it doubles as the training probe. A second copy of the pool has the idle lead-in before the demonstrator's first movement removed (the trimmed data).

\subsection{Models and training}\label{sec:3.3}

All models were trained with LeRobot~\cite{cadene2024lerobot} at commit 8352ca62 (v0.6.2), using its default recipe for each policy. We call one trained model, a combination of model line, budget and seed, a training cell.

\begin{table}[tbp]
\centering
\small
\caption{Training configurations. A training cell is one combination of model line, budget and seed.}
\label{tab:training}
\begin{tabularx}{\linewidth}{@{}>{\raggedright\arraybackslash}p{4.4cm}YYYY@{}}
\toprule
\textbf{Model line} & \textbf{Training steps} & \textbf{Checkpoints saved} & \textbf{Seeds per budget} & \textbf{Cells in the bench analysis} \\
\midrule
ACT & 100,000 & every 10,000 & 4 & 20 \\
Diffusion Policy, untrimmed data & 100,000 & every 10,000 & 4 trained, 2 benched & 10 (+2 EMA) \\
Diffusion Policy, trimmed data & 100,000 & every 10,000 & 2 (seeds 1, 3) & 10 (+4 EMA) \\
SmolVLA & 20,000 & every 2,000 & 4 & 20 \\
$\pi_{0.5}$, action expert only & 6,000 & 3,000 and 6,000 & 1 plain + 2 EMA & 15 (5 plain, 10 EMA) \\
\bottomrule
\end{tabularx}
\end{table}

ACT trains at a constant learning rate; Diffusion Policy decays its learning rate to zero at the last step on a cosine schedule, and SmolVLA on a cosine schedule compressed to its run length. This matters when reading the end of a training curve (Section~\ref{sec:6}).

\subsection{Bench protocol}\label{sec:3.4}

Each checkpoint was rolled out from a rest pose at ten standard ball placements, N00 to N09, in the standard bench lighting of~\cite{cowling2026scaling}: the room's ordinary lighting of 117 to 394 lux, dimmer than the roughly 500 lux under which the demonstrations were recorded. Each rollout had a 30-second limit and was labelled with one outcome: success, never started (the arm did not leave its rest pose), never located the ball, grip failed, or another failure. As in~\cite{cowling2026scaling}, late starts and rollouts without a grasp attempt count as other failures. The primary analysis set contains every rollout in this lighting at placements N00 to N09 from the main grid and the repeat, control and weight-comparison sessions of~\cite{cowling2026scaling}: 82 checkpoints and 930 of the 1,340 rollouts reported there. Excluded are a probe in brighter light, an earlier $\pi_{0.5}$ LoRA grid, and two trimmed-data checkpoints trained with EMA but rolled out with plain weights; the last two have no matching validation scores. We use the audited outcome labels of~\cite{cowling2026scaling}, in which 27 of the 1,340 labels were corrected from the camera recordings.

\section{Scoring method}\label{sec:4}

\subsection{How a checkpoint is scored}\label{sec:4.1}

Each checkpoint is loaded with the normalisation saved alongside it and run in evaluation mode, without gradients or image augmentation (ACT skips its VAE encoder; Diffusion Policy uses its centre crop). Frames are batched one episode at a time, so every number also exists per episode and its uncertainty is a standard error across episodes. For each checkpoint we record, on the hold-out and on the training probe:

\begin{itemize}
\item \textbf{loss}: the policy's own objective, as in training; for the three generative policies averaged over four draws of noise and time step with a fixed seed, so every checkpoint sees the same draws;
\item \textbf{action error}: the mean absolute error of Section~\ref{sec:2.3}, overall and per joint.
\end{itemize}

Final checkpoints of ACT, Diffusion Policy and SmolVLA were scored on every frame (13,372 hold-out frames, 6,139 probe frames). $\pi_{0.5}$ was scored on every fifth frame in bfloat16, since one checkpoint takes most of an hour. The 700 intermediate checkpoints were also scored on every fifth frame; on the 106 checkpoints scored both ways, this changed loss by a median of 0.8 to 3.5\% and action error by 0.2 to 0.3\%, and left the ranking of checkpoints unchanged (rank correlation 0.988 or higher). Plain and EMA weights are scored separately. The scorer refuses to run if a hold-out episode appears in a model's training set.

\subsection{Statistics}\label{sec:4.2}

\begin{itemize}
\item \textbf{Rank correlation} (Spearman's $\rho$) between an offline measure and bench success across a model's checkpoints, with p-values from 20,000 permutations. Negative means lower error goes with higher success.
\item \textbf{Expected from episodes alone.} Success is simulated 5,000 times with each checkpoint's success probability set to its budget's pooled rate, so the offline measure carries nothing beyond the budget; the observed $\rho$ is compared with the simulated range.
\item \textbf{Within-budget correlation.} Both variables are centred on their budget means and $\rho$ recomputed, with 10,000 permutations within budgets. This asks whether the measure knows which seed did better.
\item \textbf{Success rates} are pooled counts with 95\% Wilson intervals; paired counts (EMA, trimming) use Fisher's exact test; seed differences use a chi-square summed over budgets with a simulated p-value.
\end{itemize}

\subsection{Compute}\label{sec:4.3}

Final checkpoints were scored on a single workstation GPU, taking about three minutes per ACT checkpoint and under an hour per $\pi_{0.5}$ checkpoint. Intermediate checkpoints were scored on NVIDIA A10G GPUs through Hugging Face Jobs.

\section{Final checkpoints and bench success}\label{sec:5}

\subsection{Both measures improve with data}\label{sec:5.1}

\begin{figure}[tbp]
\centering
\includegraphics[width=\linewidth,height=0.42\textheight,keepaspectratio]{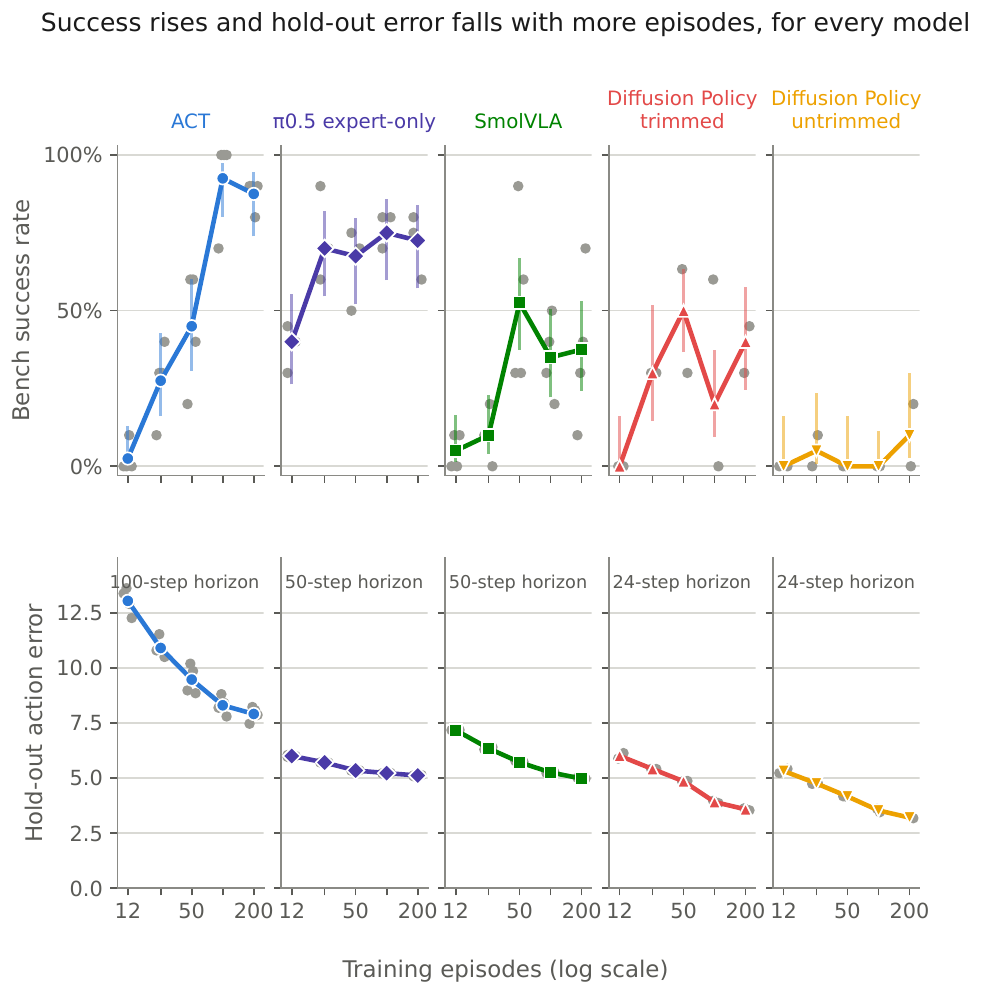}
\caption{Bench success (top) and hold-out action error (bottom) by number of training episodes. Grey dots are single checkpoints; bars are 95\% Wilson intervals on the pooled rate. Error horizons differ between columns, so compare shapes, not levels.}
\label{fig:1}
\end{figure}

For every model, bench success rises and hold-out action error falls from 12 to 200 episodes, with diminishing returns (Figure~\ref{fig:1}). ACT goes from 1 success in 40 rollouts at 12 episodes to 37 of 40 at 100. The pretrained models start higher: $\pi_{0.5}$ succeeds in 16 of 40 rollouts with 12 episodes. The gap between seen and unseen episodes narrows as data grows, and loss on the training probe rises, because a larger training set is harder to fit exactly.

\subsection{Within a model, validation adds nothing to the episode count}\label{sec:5.2}

Across each model's checkpoints, lower hold-out error goes with higher success. But the correlation is no stronger than the episode count alone would produce, and among checkpoints trained on the same episodes it vanishes.

\begin{table}[tbp]
\centering
\small
\caption{Rank correlation between hold-out action error and bench success across each model line's checkpoints, the range the episode count alone would produce, and the correlation among checkpoints trained on the same number of episodes (with its permutation p-value).}
\label{tab:within}
\begin{tabularx}{\linewidth}{@{}>{\raggedright\arraybackslash}p{3.6cm}YYYY@{}}
\toprule
\textbf{Model line} & \textbf{Checkpoints} & \textbf{$\rho$, hold-out error vs success} & \textbf{Expected from episodes alone [95\% range]} & \textbf{$\rho$ within budget (p)} \\
\midrule
ACT & 20 & $-0.88$ & $-0.89$ [$-0.95$, $-0.82$] & +0.28 (0.82) \\
$\pi_{0.5}$ expert-only & 15 & $-0.55$ & $-0.50$ [$-0.80$, $-0.13$] & $-0.22$ (0.85) \\
SmolVLA & 20 & $-0.68$ & $-0.62$ [$-0.78$, $-0.42$] & $-0.08$ (0.69) \\
Diffusion Policy, trimmed & 10 & $-0.48$ & $-0.48$ [$-0.76$, $-0.16$] & +0.11 (0.75) \\
Diffusion Policy, untrimmed & 10 & $-0.23$ & $-0.34$ [$-0.70$, +0.29] & $-0.16$ (1.00) \\
\bottomrule
\end{tabularx}
\end{table}

\begin{figure}[tbp]
\centering
\includegraphics[width=\linewidth,height=0.42\textheight,keepaspectratio]{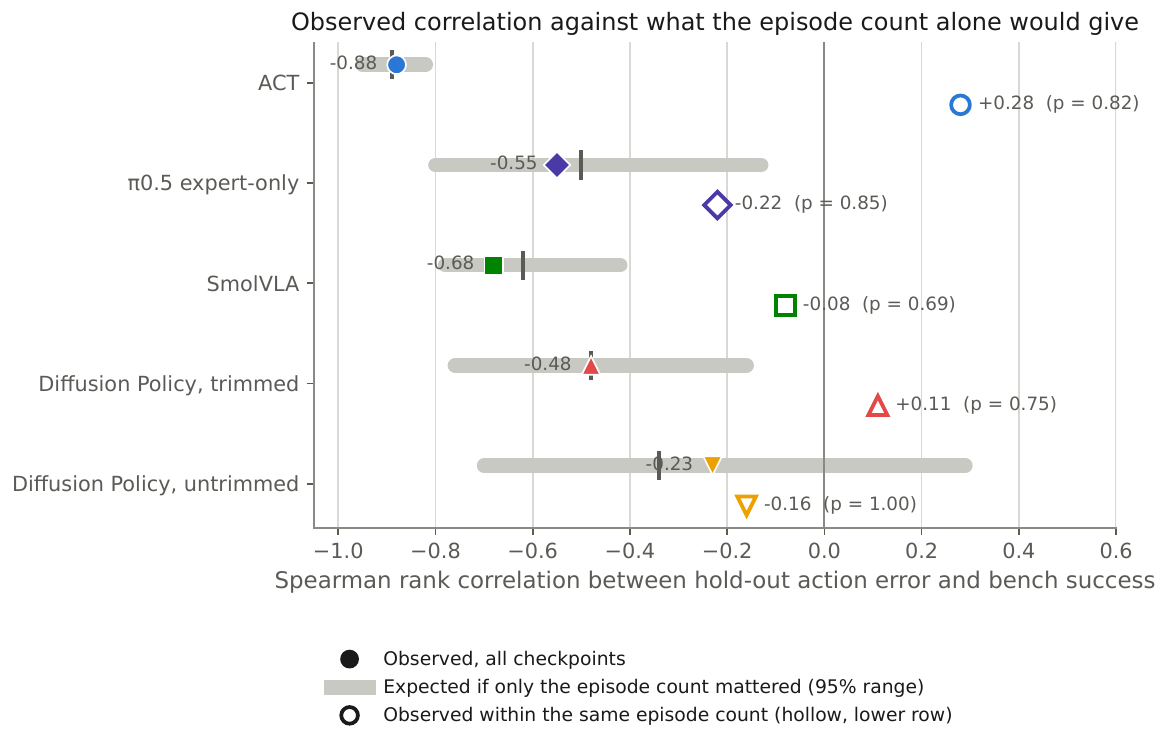}
\caption{Observed correlations (filled) against the range the episode count alone would produce (grey), and the correlation among checkpoints trained on the same number of episodes (hollow).}
\label{fig:2}
\end{figure}

Hold-out action error correlates with the logarithm of the episode count at $-0.95$ to $-0.99$ for every model. It is close to a relabelling of how much data the model saw (Figure~\ref{fig:2}).

\subsection{Across models, validation points the wrong way}\label{sec:5.3}

At 200 episodes the ranking of the five model lines by hold-out action error and the ranking by success correlate at +0.90: the lower the error, the worse the bench result. Untrimmed Diffusion Policy has the lowest hold-out error in the study and succeeded in 3 of 110 rollouts. Part of this is the horizon confound (ACT is scored over 100 steps, Diffusion Policy over 24), but the inversion survives within the one like-for-like pair (Section~\ref{sec:5.5}).

\begin{figure}[tbp]
\centering
\includegraphics[width=\linewidth,height=0.42\textheight,keepaspectratio]{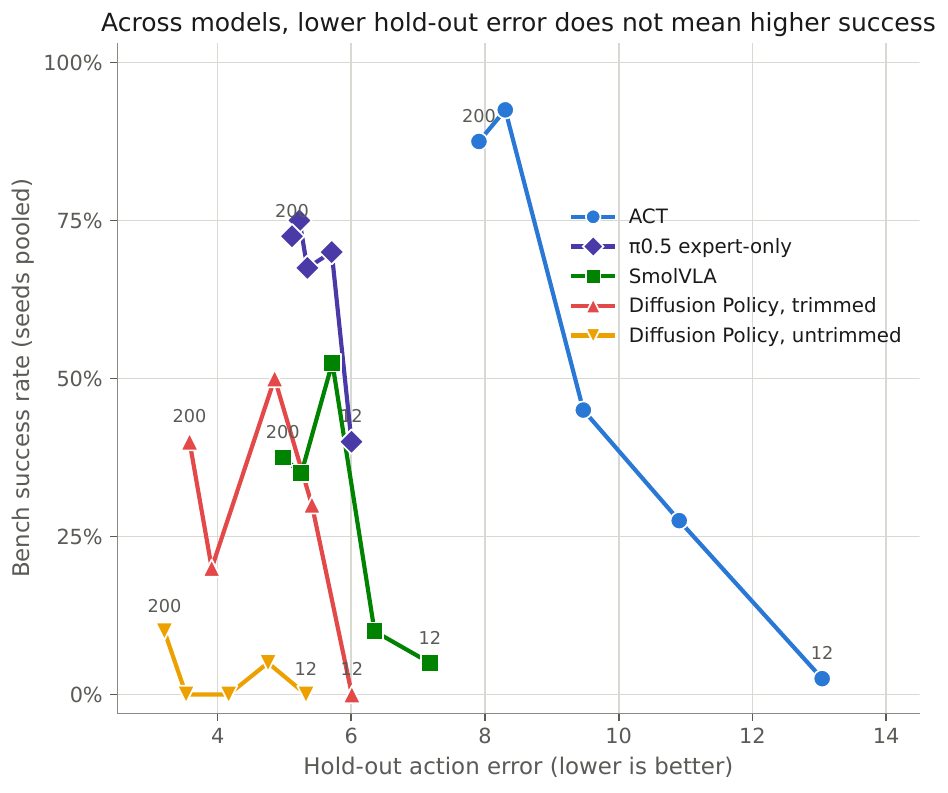}
\caption{Hold-out action error against bench success for each model line from 12 to 200 episodes. If validation predicted success across models, the lines would run from top left to bottom right.}
\label{fig:3}
\end{figure}

\subsection{Initiation failure is invisible to validation}\label{sec:5.4}

\begin{figure}[tbp]
\centering
\includegraphics[width=\linewidth,height=0.42\textheight,keepaspectratio]{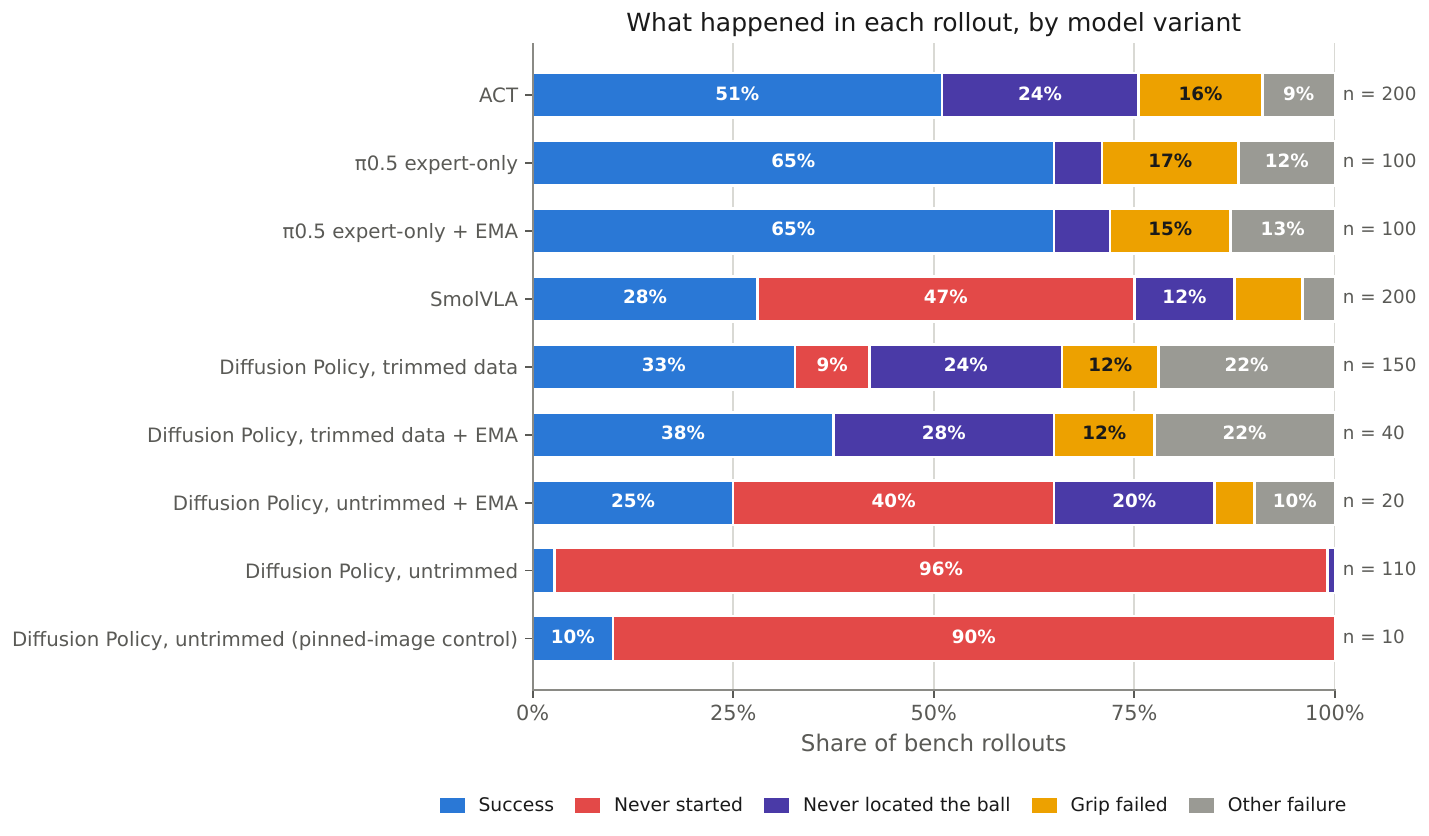}
\caption{Outcome of every bench rollout by model variant (standard dim lighting, placements N00 to N09). As in~\cite{cowling2026scaling}, late starts and rollouts without a grasp attempt count as other failures.}
\label{fig:4}
\end{figure}

The most common failure in the study is a policy that never moves the arm from its rest pose: 96\% of untrimmed Diffusion Policy rollouts, 47\% of SmolVLA rollouts, 9\% of trimmed Diffusion Policy rollouts, and none for ACT or $\pi_{0.5}$ (Figure~\ref{fig:4}). A hold-out made of demonstrations cannot contain this situation, because the demonstrator always moved.

\subsection{Once the arm starts, validation is informative}\label{sec:5.5}

For SmolVLA, hold-out action error is only weakly related to how often the arm starts ($\rho$ = $-0.18$) but strongly related to success among rollouts that did start ($\rho$ = $-0.81$). SmolVLA and $\pi_{0.5}$ are the one pair scored over the same 50-step horizon. At 200 episodes their hold-out errors are almost equal (4.98 and 5.11), and so are their success rates among started rollouts (71\% and 73\%). The difference in overall success, 38\% against 73\%, is almost all initiation.

\begin{figure}[tbp]
\centering
\includegraphics[width=\linewidth,height=0.42\textheight,keepaspectratio]{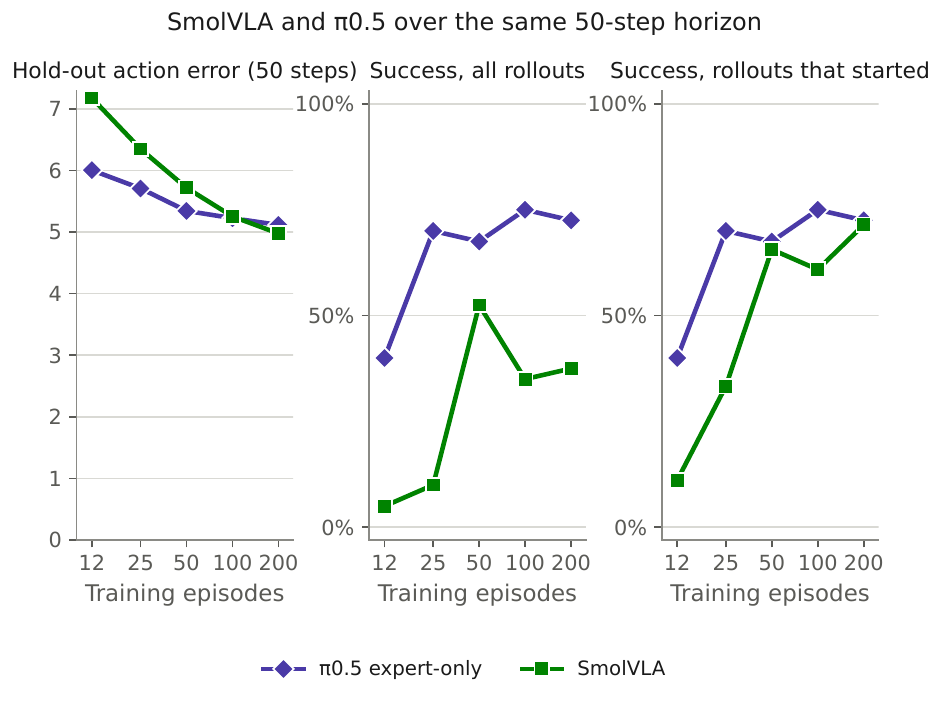}
\caption{SmolVLA and $\pi_{0.5}$ by budget: hold-out action error (left), success over all rollouts (middle), and success over the rollouts in which the arm started (right).}
\label{fig:5}
\end{figure}

\subsection{EMA and trimming change the bench, not the validation}\label{sec:5.6}

\begin{figure}[tbp]
\centering
\includegraphics[width=\linewidth,height=0.42\textheight,keepaspectratio]{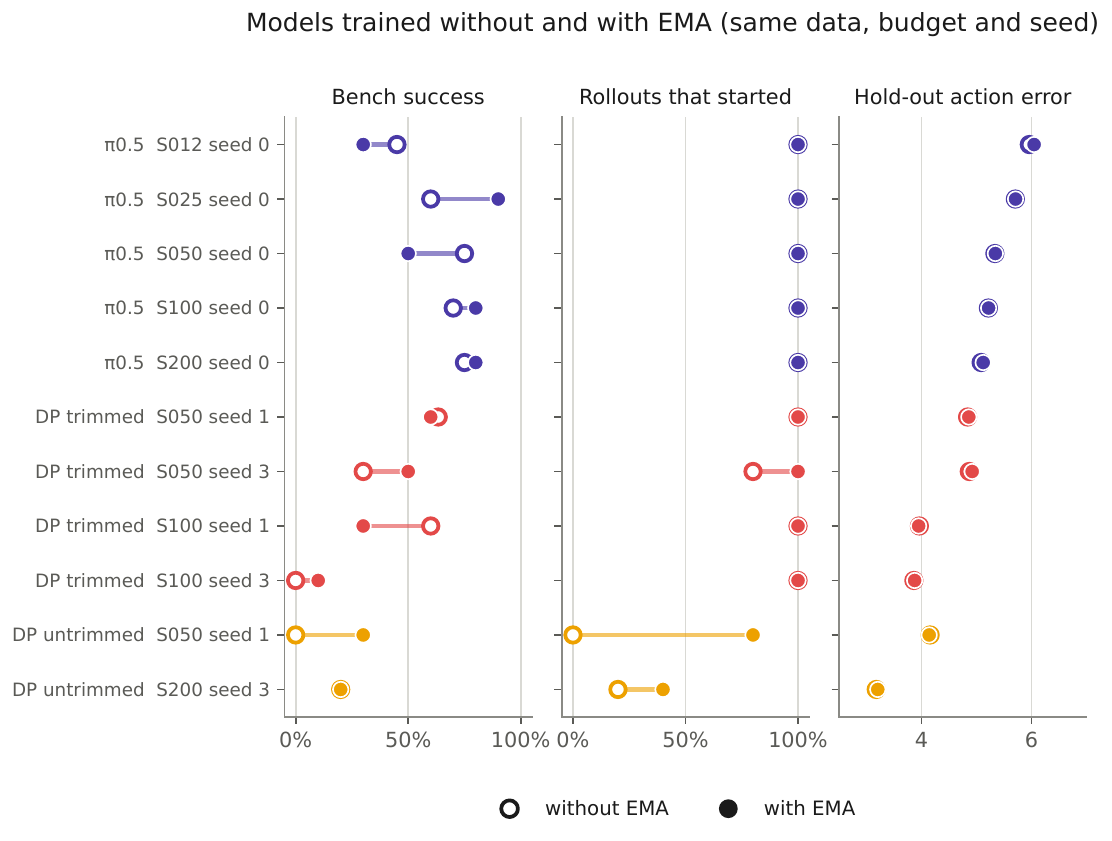}
\caption{Each row pairs a model trained without EMA (hollow) with a model trained separately with EMA on the same data, budget and seed (filled), scored with its averaged weights.}
\label{fig:6}
\end{figure}

We compare 11 pairs of models trained on the same data, budget and seed, one without and one with an exponential moving average (EMA) of the weights; the two members of each pair were trained separately. Their hold-out action error differs by at most 1.5\% and their hold-out loss by at most 3.0\%. On the bench, $\pi_{0.5}$ and trimmed Diffusion Policy are unchanged, but the two untrimmed Diffusion Policy models trained with EMA started in 12 of 20 rollouts, against 2 of 20 for their counterparts trained without it (Fisher p = 0.002; Figure~\ref{fig:6}). The EMA models were retrained and benched on different days, so with two pairs the difference cannot be attributed to EMA rather than to variation between training runs~\cite{cowling2026scaling}. Either way, it is a large difference in behaviour on the robot that validation did not register.

\begin{figure}[tbp]
\centering
\includegraphics[width=\linewidth,height=0.42\textheight,keepaspectratio]{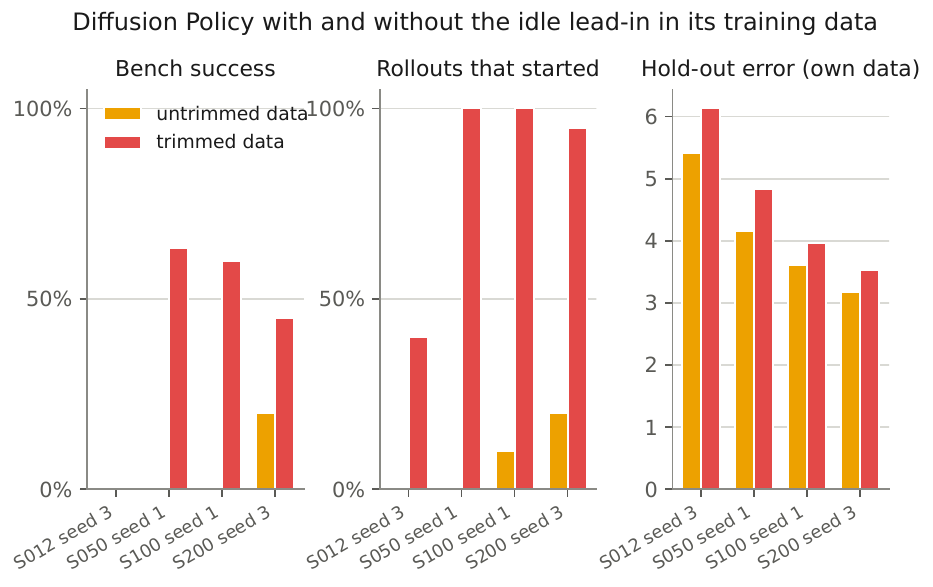}
\caption{The same Diffusion Policy cells trained with and without the idle lead-in.}
\label{fig:7}
\end{figure}

Trimming the idle lead-in from the training data took four paired Diffusion Policy cells from 2 successes in 40 rollouts to 34 in 70. Scored on their own validation data, the trimmed models looked slightly worse (Figure~\ref{fig:7}). The trimmed and untrimmed hold-outs contain different frames, so the two sets of numbers are not directly comparable, but nothing in either predicted success rising from 5\% to 49\%.

\subsection{Seeds differ on the bench, not in validation}\label{sec:5.7}

SmolVLA seeds trained on the same episodes differ on the bench by more than chance (simulated p = 0.04), mostly in how often the arm starts: at 200 episodes the four seeds started 2, 3, 6 and 10 of their ten rollouts. Their hold-out action error differs by about 1\%. ACT's seed differences on the bench are within chance (p = 0.11), and its hold-out error varies by 4 to 7\% between seeds.

\subsection{Loss on the training episodes orders the budgets backwards}\label{sec:5.8}

A model trained on 12 episodes fits them almost perfectly, so its probe loss is the lowest in the study while its bench success is the lowest. Probe loss therefore correlates positively with success: +0.86 for ACT, +0.67 for SmolVLA, +0.51 for $\pi_{0.5}$. Training loss measures how closely a model fits its own demonstrations, not how well it performs the task, and cannot be used to compare data budgets.

\section{Training dynamics}\label{sec:6}

We scored all 700 saved checkpoints of the 70 ACT, Diffusion Policy and SmolVLA runs, ten per run, on the hold-out and on the training probe. Loss on the training probe falls to the last checkpoint for every model and budget. This reflects fitting of the training set and does not show whether a model was trained long enough; the hold-out curves do.

\subsection{For the generative policies, loss and action error disagree}\label{sec:6.1}

\begin{figure}[tbp]
\centering
\includegraphics[width=\linewidth,height=0.42\textheight,keepaspectratio]{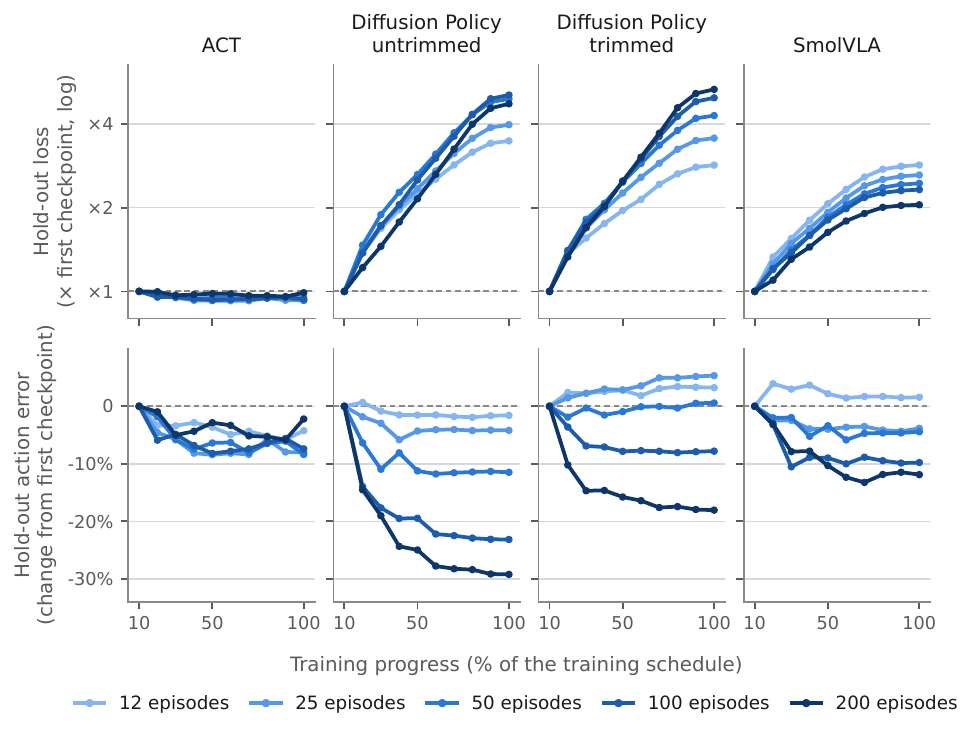}
\caption{Hold-out loss as a multiple of its first-checkpoint value (top, log scale) and hold-out action error as a change from its first-checkpoint value (bottom), against training progress. Shades of blue are the episode budgets.}
\label{fig:8}
\end{figure}

For ACT, hold-out loss and hold-out action error tell the same story: both drift down a few percent early and then stay flat. For Diffusion Policy and SmolVLA they tell opposite stories (Figure~\ref{fig:8}). Hold-out loss is lowest at the first saved checkpoint for all 15 of their model and budget combinations and climbs steadily, to 3.5 to 5.1 times its starting value for untrimmed Diffusion Policy, 2.8 to 5.3 for trimmed and 2.0 to 2.9 for SmolVLA. Over the same steps, the error of the actions they produce stays level at small budgets and falls at large ones, by up to 29\% for untrimmed Diffusion Policy and 12\% for SmolVLA at 200 episodes.

The explanation lies in what the loss measures (Section~\ref{sec:2.2}). These models are trained to repair a corrupted demonstration at every level of corruption. Trained long on a fixed set of demonstrations, the network learns to pull any noisy input firmly towards the actions it has seen~\cite{he2025demystifying}. On a hold-out frame the correct action is not one of those, so at many noise levels the repair is confidently wrong and the averaged loss rises. The action the policy outputs comes from running the whole denoising chain, which converges on a training-like action that fits the situation increasingly well. Action error scores that output.

Consequently, early stopping on hold-out loss would have halted every Diffusion Policy and SmolVLA run at 10\% of its schedule, where hold-out action error was up to 41\% higher than at the end of training. For ACT, whose loss is itself an action error, the two criteria agree.

\subsection{Convergence on unseen demonstrations}\label{sec:6.2}

On unseen demonstrations, only Diffusion Policy at 100 and 200 episodes was still improving when training stopped.

\begin{figure}[tbp]
\centering
\includegraphics[width=\linewidth,height=0.42\textheight,keepaspectratio]{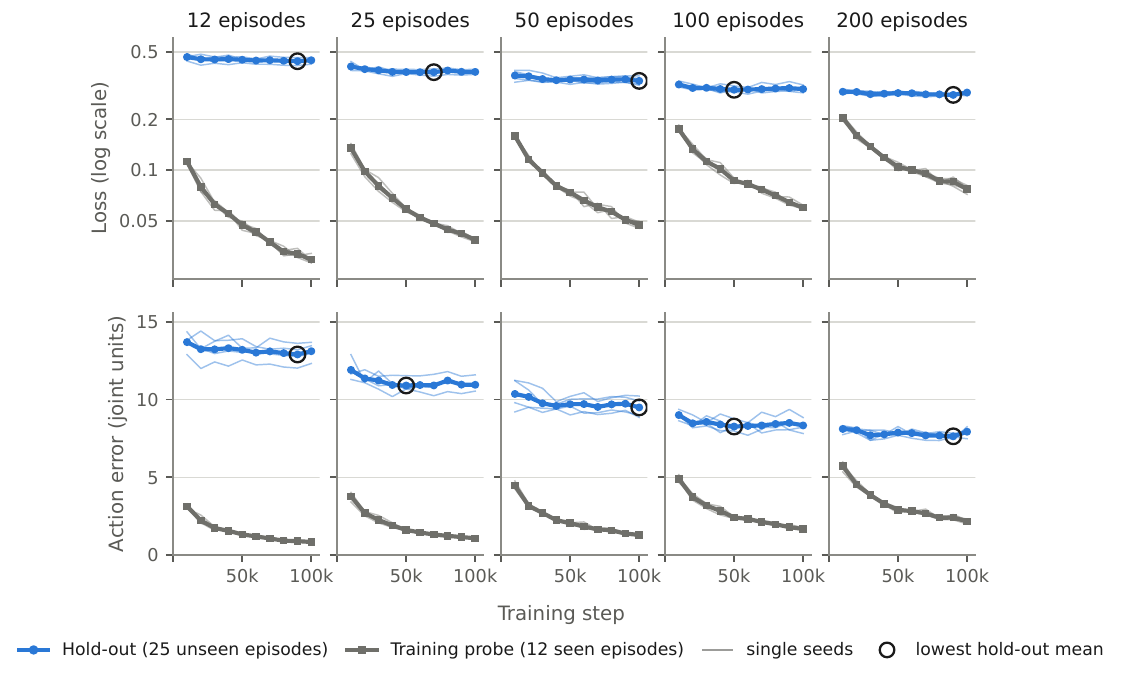}
\caption{ACT: hold-out (blue) and training-probe (grey) loss (top, log scale) and action error (bottom) at every budget. Thin lines are seeds, bold lines their mean; the ring marks the lowest hold-out mean.}
\label{fig:9}
\end{figure}

\begin{figure}[tbp]
\centering
\includegraphics[width=\linewidth,height=0.42\textheight,keepaspectratio]{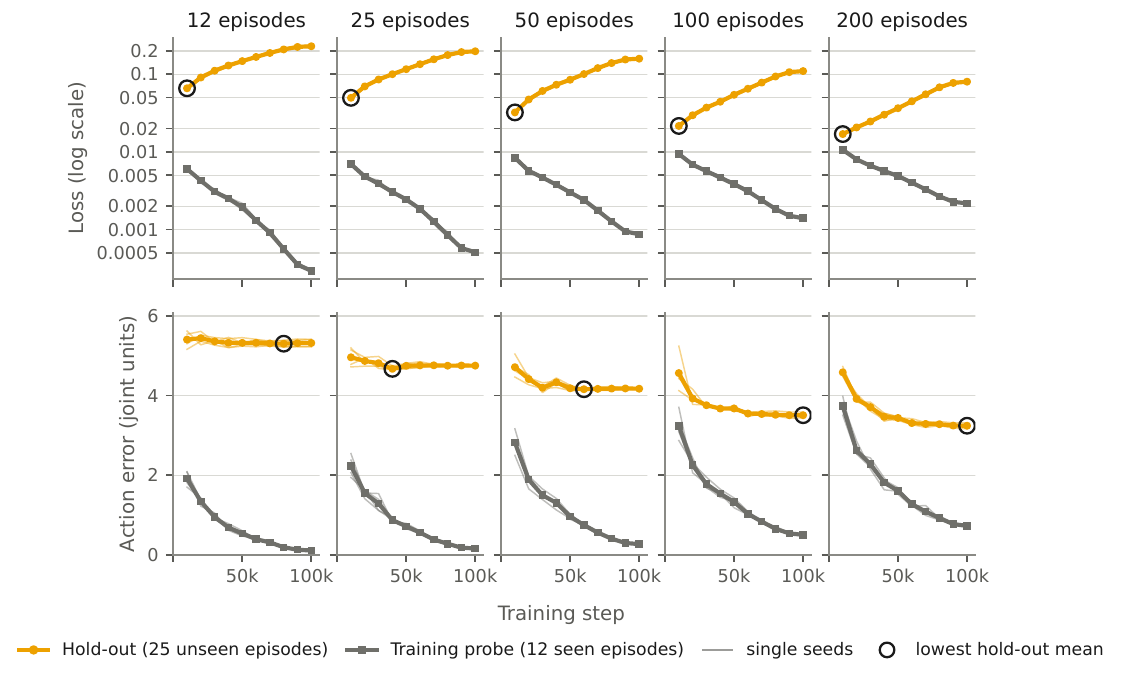}
\caption{Diffusion Policy, untrimmed data, in the same layout. Hold-out loss rises from the first checkpoint while hold-out action error falls.}
\label{fig:10}
\end{figure}

\begin{figure}[tbp]
\centering
\includegraphics[width=\linewidth,height=0.42\textheight,keepaspectratio]{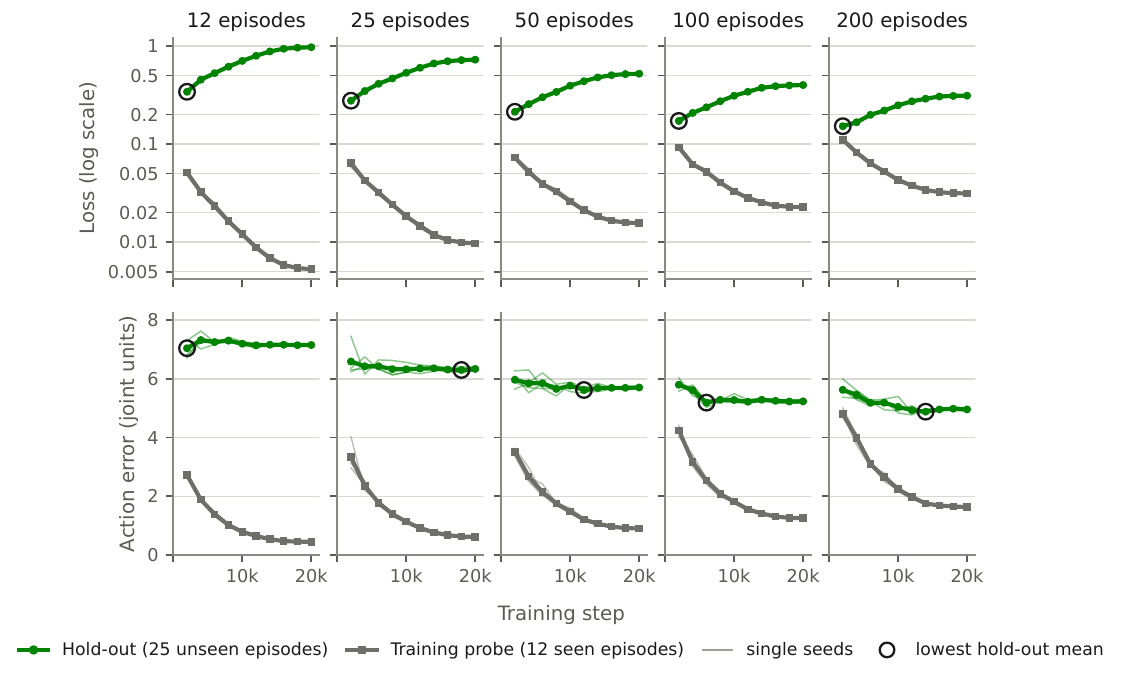}
\caption{SmolVLA in the same layout (20,000 training steps).}
\label{fig:11}
\end{figure}

\begin{itemize}
\item \textbf{ACT} reaches within 2\% of its final hold-out action error by 20,000 to 40,000 steps at every budget. Over the second half of training the change is $-2.2$\% to +0.9\%, inside the 4 to 7\% spread between seeds. ACT trains at a constant learning rate, so this plateau is the model's own (Figure~\ref{fig:9}).
\item \textbf{Diffusion Policy} at 12 to 50 episodes is within 2\% of its final value by 10,000 to 30,000 steps. At 100 and 200 episodes (untrimmed) its hold-out action error fell 4.6\% and 5.7\% over the second half, in all four seeds; a straight-line fit per seed gives p = 0.048 and 0.001. Trimmed Diffusion Policy at 200 episodes points the same way ($-2.7$\%, p = 0.10 with two seeds). No other combination of model and budget shows a trend (p $\geq$ 0.10). Its cosine schedule drives the learning rate to zero at the end, so the remaining gain may be larger than the curves suggest (Figure~\ref{fig:10}).
\item \textbf{SmolVLA} starts from pretrained weights and reaches within 2\% of its final hold-out action error by 2,000 to 10,000 steps; over the second half it changes by $-1.6$\% to +0.2\% (Figure~\ref{fig:11}).
\end{itemize}

\subsection{More data versus more training steps}\label{sec:6.3}

\begin{figure}[tbp]
\centering
\includegraphics[width=\linewidth,height=0.42\textheight,keepaspectratio]{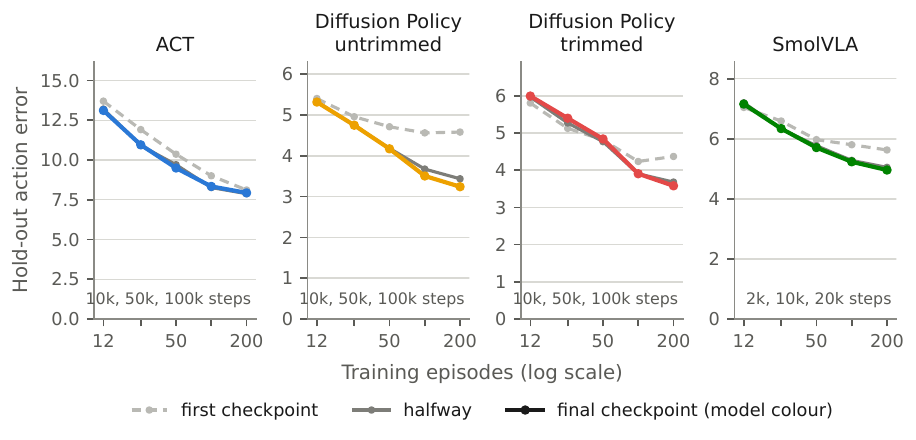}
\caption{Hold-out action error by episode count at the first, middle and final checkpoint. The vertical gap between lines is what training longer bought; the slope is what more data bought.}
\label{fig:12}
\end{figure}

Going from 12 to 200 episodes reduced hold-out action error by 31 to 40\% for every model. Training ten times longer reduced it by 2\% for ACT and, at 12 episodes, by at most 4\% for any model. For Diffusion Policy and SmolVLA the gain from more steps grows with the budget, to 29\% for untrimmed Diffusion Policy at 200 episodes: more demonstrations also take more steps to absorb (Figure~\ref{fig:12}).

\begin{table}[tbp]
\centering
\small
\caption{Change in hold-out action error from more data and from more training steps, and the change in hold-out loss over training.}
\label{tab:datasteps}
\begin{tabularx}{\linewidth}{@{}>{\raggedright\arraybackslash}p{3.6cm}YYYY@{}}
\toprule
\textbf{Model line} & \textbf{12 $\rightarrow$ 200 episodes (final step)} & \textbf{10$\times$ more steps at 12 episodes} & \textbf{10$\times$ more steps at 200 episodes} & \textbf{Hold-out loss over training, 12 / 200 episodes} \\
\midrule
ACT & $-40$\% & $-4$\% & $-2$\% & $\times$1.0 / $\times$1.0 \\
Diffusion Policy, untrimmed & $-39$\% & $-2$\% & $-29$\% & $\times$3.5 / $\times$4.7 \\
Diffusion Policy, trimmed & $-40$\% & +3\% & $-18$\% & $\times$2.8 / $\times$5.3 \\
SmolVLA & $-31$\% & +2\% & $-12$\% & $\times$2.8 / $\times$2.0 \\
\bottomrule
\end{tabularx}
\end{table}

\subsection{Bench results come from near-best checkpoints}\label{sec:6.4}

Every bench result comes from a final checkpoint. For all 20 combinations of model line and budget, the final checkpoint is within 5.3\% of the best hold-out action error reached during training (median 0.8\%), against differences of 31 to 40\% between 12 and 200 episodes. Selecting checkpoints by validation would therefore not have changed the scaling results of~\cite{cowling2026scaling}. Whether bench success also stayed flat while the offline curves did is untested: in~\cite{barannikov2024validation}, ACT's simulated success rose by more than 15\% while its validation loss changed by 2\%.

\section{Spatial coverage of ball placements}\label{sec:7}

\subsection{Recovering positions from the front camera}\label{sec:7.1}

We located all three balls at the start of every recorded episode and of one bench session, in millimetres on the board. In each frame the 50 mm grid is detected and fitted, which gives the camera's pose; each ball is then fitted as a sphere of known size seen by that camera, so balls cut off by the frame edge or partly hidden can still be placed.

The camera model is consistent across all 225 episodes: the front camera sits 82 mm above the board, tilted 28$^\circ$ down, with a focal length of about 735 px. Fitted ball diameters come out at 29 mm (green) and 31 mm (red) with a spread under 1.5 mm, an independent check on the scale. Positions from frame 0 and frame 30 of the same episode agree to 0.7 mm (median; 7 mm at most). Hiding parts of fully visible balls on purpose shows that a ball with only 30\% of its width in view is still placed to within 1.1 mm (median; 3.4 mm at the 90th percentile). In one episode the green ball was placed after recording began, and frame 30 was used. Relative to the grid, the camera shifted by about 5 mm between recording batches 1 to 4 and 5 to 9; positions are reported relative to the camera.

\subsection{Coverage grows with the budget}\label{sec:7.2}

\begin{figure}[tbp]
\centering
\includegraphics[width=\linewidth,height=0.42\textheight,keepaspectratio]{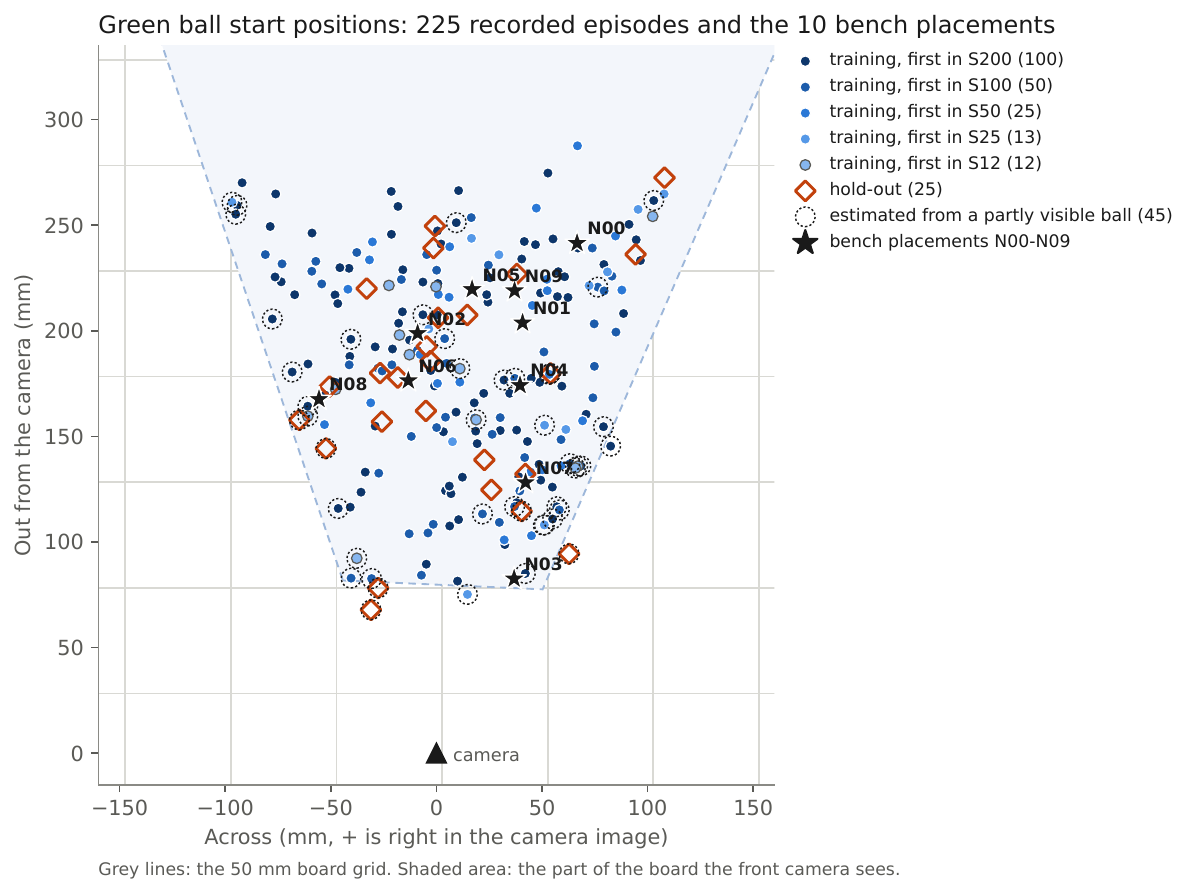}
\caption{Green ball start positions in all 225 recorded episodes, coloured by the smallest budget that includes them, with the hold-out (diamonds) and the ten bench placements (stars). Dotted rings mark balls placed from a partial view.}
\label{fig:13}
\end{figure}

\begin{figure}[tbp]
\centering
\includegraphics[width=\linewidth,height=0.42\textheight,keepaspectratio]{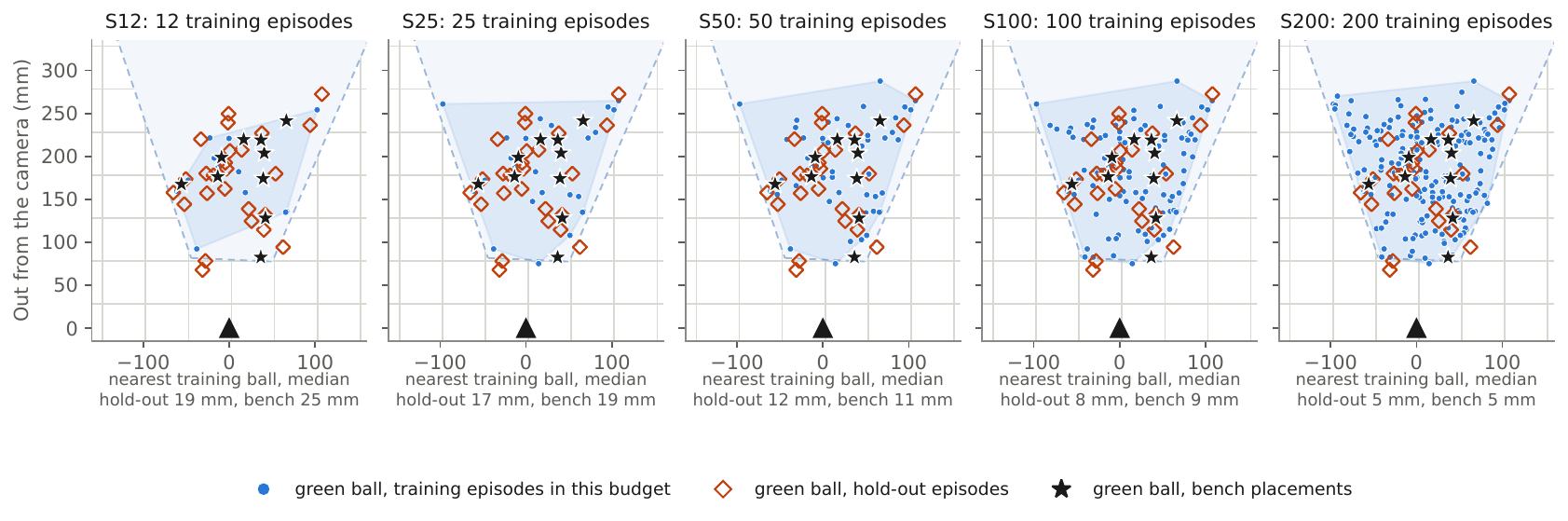}
\caption{Training green balls in each budget (dots, shaded by their area), with hold-out balls (diamonds) and bench placements (stars).}
\label{fig:14}
\end{figure}

Green start positions cover the whole floor the camera sees, about 200 mm across and 70 to 290 mm out from the camera (Figure~\ref{fig:13}). How close a hold-out ball is to the nearest training ball shrinks steadily with data: a median of 19 mm at 12 episodes, 12 mm at 50 and 5 mm at 200 (Figure~\ref{fig:14}). With 12 episodes, 9 of the 25 hold-out balls lie outside the area spanned by the training balls; from 50 episodes on, 5 do, all at the edges of the camera's view. This is the geometric side of Section~\ref{sec:5.1}: hold-out error falls as the training set covers the hold-out's situations more densely.

\subsection{The bench placements}\label{sec:7.3}

The ten bench placements were mapped from a single bench session (29 September 2026), chosen because the ball positions logged for it were closest to the typical position of each placement across sessions. The bench camera pose matches that of the recordings (82 mm high, tilted 27$^\circ$ down, rotated about 2.5$^\circ$ relative to them). Nine of the ten bench green balls lie inside the area spanned by the training balls at every budget. N03, at the near edge of the view and 82 mm out, lies just outside it even at 200 episodes, although a training ball is 6 mm away. At 200 episodes every bench green ball is within 11 mm of a training ball (median 5 mm); at 12 episodes the median is 25 mm.

\begin{figure}[tbp]
\centering
\includegraphics[width=\linewidth,height=0.42\textheight,keepaspectratio]{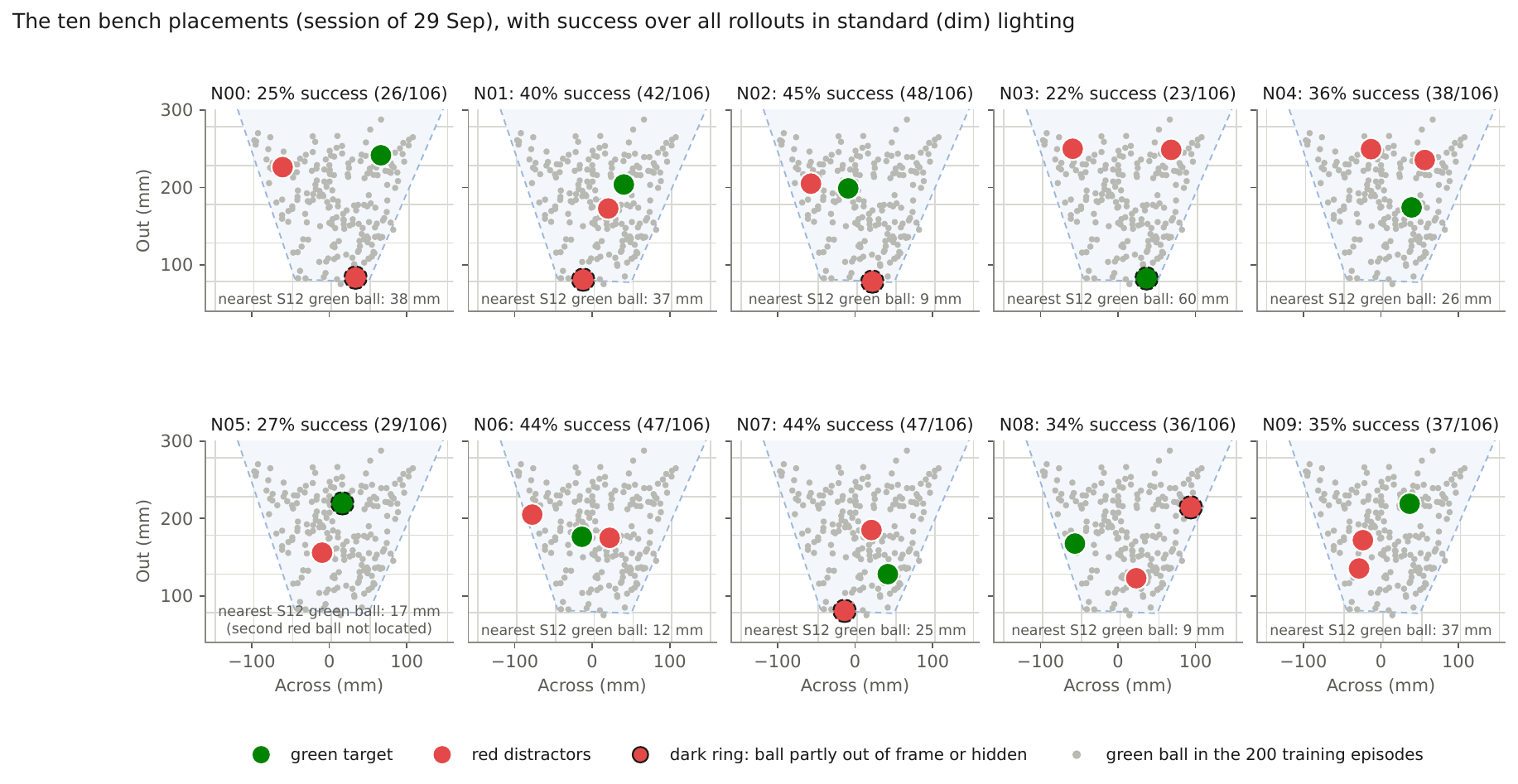}
\caption{The ten bench layouts over the 200 training positions, with each placement's success rate over all rollouts in standard (dim) lighting (106 per placement).}
\label{fig:15}
\end{figure}

\subsection{Distance from the training data and bench success}\label{sec:7.4}

The two hardest placements are the two furthest from what the 12-episode models saw: N03 (22\% success) is 60 mm from the nearest S12 green ball and N00 (25\%) 38 mm. The easiest, N02 (45\%), is 9 mm away. Across the ten placements, success and distance at S12 correlate at $\rho$ = $-0.63$ (p = 0.053).

\begin{figure}[tbp]
\centering
\includegraphics[width=\linewidth,height=0.42\textheight,keepaspectratio]{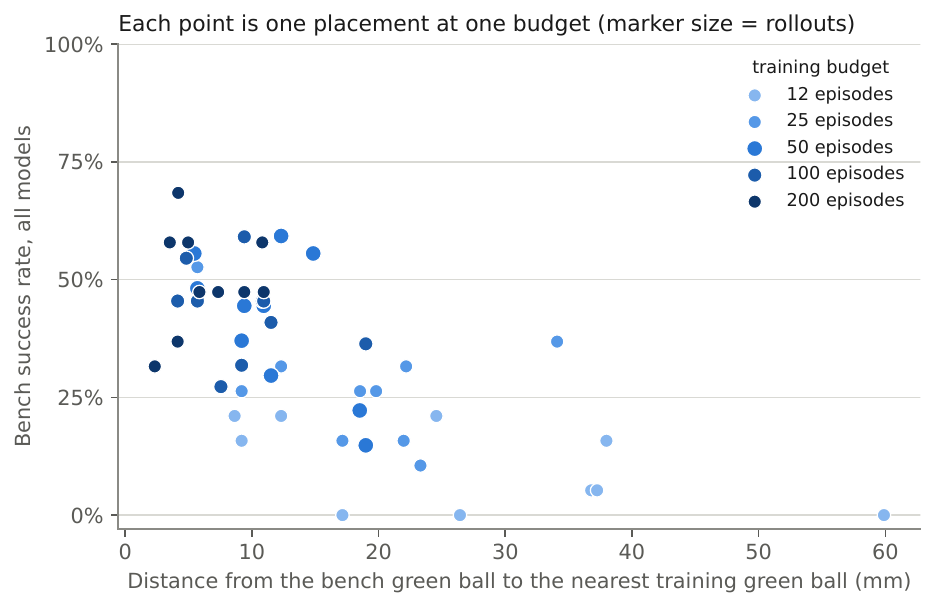}
\caption{Bench success rate of each placement at each budget against the distance from its green ball to the nearest training green ball. Marker size is the number of rollouts.}
\label{fig:16}
\end{figure}

Pooled over budgets, placements far from any training ball succeed less often ($\rho$ = $-0.65$ over 50 placement-by-budget cells), but distance and success both follow the budget. Within each budget the relation is weaker and not significant (mean $\rho$ = $-0.26$, permutation p = 0.08). It is clearest at 12 episodes ($\rho$ = $-0.53$) and absent at 200 ($\rho$ = +0.15), once every placement has a training example within about 10 mm (Figure~\ref{fig:16}). With ten placements, distance is entangled with every other feature of a layout, such as N03's green ball being partly out of view, so the association is suggestive rather than established.

\section{Discussion}\label{sec:8}

\subsection{What the validation numbers measure}\label{sec:8.1}

Read on their own, the validation results are internally consistent: every model learns more from more demonstrations, with diminishing returns; pretrained models generalise from fewer examples; ACT stops improving on unseen data early in its schedule; and models trained with and without EMA score almost identically. The placement maps show the mechanism: the situations in the hold-out are covered more densely as the training set grows. Validation is a clean readout of how much a model has learned from its data.

What it does not measure is whether the policy, run closed loop from a rest pose, will do the task. The study shows that gap in four independent ways: the ranking across models is inverted, seeds with indistinguishable validation numbers differ on the bench, EMA and trimming changed the bench without moving validation, and for generative policies the loss itself moves the wrong way during training.

\subsection{Why validation cannot see initiation failure}\label{sec:8.2}

The failure that separates SmolVLA and Diffusion Policy from ACT and $\pi_{0.5}$, and SmolVLA's seeds from one another, is the arm never starting. A hold-out drawn from demonstrations cannot contain it, because the demonstrator always moved. The frames closest to it are the idle lead-in at the start of an untrimmed demonstration, where the correct answer is to hold still, which is exactly what a policy that never starts does. An idle-prone policy is rewarded by those frames and penalised by none. The mechanism resembles the inertia problem described for driving policies~\cite{codevilla2019limitations}: where the data show a stationary agent usually remaining stationary, being at rest becomes a predictor of staying at rest.

This is why trimming is the most informative single experiment. Removing the idle lead-in removed the behaviour from training and the frames that rewarded it from validation, so the trimmed models' validation numbers got slightly worse while their bench success rose from 5\% to 49\%. A validation set that could detect initiation failure would need rest-pose observations paired with the demonstrator's first movement as the target, which is the opposite of what a recording contains.

Lighting adds to the blind spot. The bench ran in the room's ordinary lighting, dimmer than the recordings. In the brighter-light probe of~\cite{cowling2026scaling}, untrimmed Diffusion Policy started in 16 of 50 rollouts, against 2 of 50 for the same checkpoints in dim light, and SmolVLA in 36 of 50 against 32 of 50. Part of the initiation failure is therefore a response to a change in conditions between recording and deployment, which a hold-out recorded under the training conditions cannot reveal either.

\subsection{Loss and action error for generative policies}\label{sec:8.3}

For diffusion and flow-matching policies, the loss scores the repair of a corrupted demonstration at every noise level, not the action the policy outputs. As these models memorise their training actions~\cite{he2025demystifying}, the repair of unseen actions gets worse while the output improves. Their hold-out loss is therefore not a measure of generalisation in the usual sense, and it inverts the usual reading of a validation curve.

Action error has limits of its own. Errors grow with the horizon, so levels differ by policy for mechanical reasons. And the sampling policies choose one of several plausible motions: on a frame where the demonstrator could reasonably have done two things, they are sometimes scored as wrong for doing the other. ACT, which regresses towards the average motion, is not penalised in that way, which may be part of why its error is the highest in the study while its success is the highest of the models trained from scratch.

\subsection{Why seeds do not separate}\label{sec:8.4}

A success rate from ten rollouts carries about $\pm$15 percentage points of uncertainty; hold-out error is averaged over 13,372 frames and is precise to a few percent. Among seeds that all start, the bench differences are mostly rollout noise. Among seeds that do not all start, the difference is initiation, which validation does not see. In both cases there is nothing for a within-budget correlation to find.

\subsection{Implications for evaluation practice}\label{sec:8.5}

These results suggest a division of labour between offline metrics and evaluation on the robot. Hold-out action error is a sound criterion for deciding when to stop training: it plateaued within 20 to 40\% of ACT's schedule, and it identified Diffusion Policy at 100 and 200 episodes as the only configurations still improving when training ended. For diffusion and flow-matching policies, hold-out loss should not be used for this purpose, because it rises from the first checkpoint while the policy's outputs improve. Whenever action error is reported, its horizon should be stated. Validation is also useful for detecting broken runs, incorrect episode lists and faulty training recipes, which appear as outliers against the rest of a grid.

On this evidence, hold-out metrics should not be used to choose between seeds, between converged checkpoints or between model families, or to decide which configurations merit robot evaluation: in this study the configuration with the lowest hold-out error performed worst on the bench. Training loss should not be used to compare data budgets, since it orders them backwards.

For bench evaluation, reporting initiation failure separately from other failure modes, and success among the rollouts in which the policy started alongside overall success, divides closed-loop performance into a component that held-out imitation metrics track and one they cannot observe. An offline test for initiation would have to be constructed deliberately (Section~\ref{sec:10}).

\section{Limitations}\label{sec:9}

All results come from one pick-and-place task on one SO-101 setup in one laboratory, and should be read as a detailed case study rather than a general law. The bench lighting was dimmer than the lighting under which the demonstrations were recorded, so bench success reflects robustness to that change as well as generalisation to new placements. Most checkpoints have ten bench rollouts, about $\pm$15 percentage points of uncertainty each; the analysis therefore relies on pooled rates and paired comparisons, and differences between single training cells should not be over-read. The grids are uneven: $\pi_{0.5}$ has two seeds (seed 3 only with EMA weights) and saved only its halfway and final checkpoints, the untrimmed Diffusion Policy line was benched with two seeds per budget, and the trimmed line has two seeds.

Action error is scored over each policy's own execution horizon (24 to 100 steps), so SmolVLA against $\pi_{0.5}$ is the only like-for-like comparison of levels. $\pi_{0.5}$ and all intermediate checkpoints were scored on every fifth frame, and $\pi_{0.5}$ in bfloat16; the effect of the stride was checked (Section~\ref{sec:4.1}), the effect of reduced precision was not. Trimmed and untrimmed models are validated on different hold-out frames, so their validation numbers are not directly comparable. The end of the Diffusion Policy and SmolVLA curves is shaped by their learning-rate schedules, taken to be LeRobot's defaults, so the remaining gain for Diffusion Policy at 100 and 200 episodes is uncertain.

Bench results exist only for final checkpoints, so the assumption that bench success follows hold-out action error along a training run is not tested. Bench ball positions were mapped from a single session, and placements shift by a few millimetres between sessions; ten placements are too few to separate distance from the training data from other features of a layout.

\section{Future work}\label{sec:10}

Several questions raised by these results can be addressed offline, with the existing recordings, checkpoints and bench record. The most direct is an offline initiation probe: presenting each policy with the opening frames of the hold-out demonstrations, while the arm is still at rest, and measuring whether its predicted actions would move the arm within the first second. If this quantity tracks each checkpoint's bench rate of starting, it would provide an offline test for the failure mode that validation cannot observe. Two refinements of action error are also open: restricting it to the approach and grasp segments, following~\cite{huang2026cimse}, and scoring all policies over a common 24-step horizon to remove the horizon confound from cross-model comparisons. Finally, Diffusion Policy at 100 and 200 demonstrations was still improving on unseen data when training stopped, and its behaviour under a longer schedule remains to be measured.

\section{Conclusion}\label{sec:11}

Validation loss and hold-out action error are good measures of learning and poor measures of performance on the robot. Across four policy families, five data budgets and 930 bench rollouts, they show cleanly that every model learns more from more demonstrations, that pretrained models generalise from fewer, that ACT and SmolVLA stopped improving on unseen data well before their schedules ended, and that only Diffusion Policy with 100 or more demonstrations was still improving.

They predict bench success no better than the episode count does, say nothing about which seed will do better, and rank models in the wrong order. The reason is a specific blind spot rather than noise: the most common failure in the study, the arm never starting, cannot appear in a hold-out made of demonstrations. Where that failure is absent, among rollouts that started, hold-out error and success agree well. For diffusion and flow-matching policies the hold-out loss is misleading even as a training diagnostic, rising from the first checkpoint while the actions they produce improve.

Hold-out action error is therefore suited to deciding when to stop training and to detecting broken runs. Choosing between seeds, checkpoints and model families still requires evaluation on the robot, which is more informative when success among the rollouts in which the policy started is reported alongside overall success.

\bibliographystyle{unsrtnat}
\bibliography{refs}

\appendix
\renewcommand{\thetable}{A\arabic{table}}
\setcounter{table}{0}
\section{Experimental details}\label{app:A}

Table~\ref{tab:details} lists the data, scoring components and settings from which every number in this paper is computed. Both copies of the demonstration dataset are publicly available on the Hugging Face Hub under the Apache 2.0 licence: \href{https://huggingface.co/datasets/fysical/greenball_pool_225_v1}{fysical/greenball\_pool\_225\_v1} and \href{https://huggingface.co/datasets/fysical/greenball_pool_225_v1_trimmed}{fysical/greenball\_pool\_225\_v1\_trimmed}.

\begin{table}[h]
\centering
\small
\caption{Data, scoring components and settings.}
\label{tab:details}
\begin{tabularx}{\linewidth}{@{}>{\raggedright\arraybackslash}p{4.3cm}YY@{}}
\toprule
\textbf{Component} & \textbf{Description} & \textbf{Settings} \\
\midrule
Demonstrations (\nolinkurl{fysical/greenball_pool_225_v1}) & 225 teleoperated demonstrations; a trimmed copy has the idle lead-in removed & 30 fps; front and wrist cameras, 640 $\times$ 480 \\
Data split (\nolinkurl{subsets.json}) & 25 hold-out episodes and nested training subsets S12 to S200 & data seed 20260922; hold-out stratified over nine recording batches \\
Scorer (\nolinkurl{val_loss.py}) & loss and action error per checkpoint, overall and per episode & evaluation mode; 4 noise draws, seed 20261007; probe S12; plain or EMA weights; rejects any model whose training set contains a hold-out episode \\
Final-checkpoint scores & 128 checkpoints, on the hold-out and the training probe & every frame ($\pi_{0.5}$ every fifth frame, bfloat16) \\
Training-curve scores & 700 intermediate checkpoints of ACT, Diffusion Policy and SmolVLA & every fifth frame; NVIDIA A10G \\
Bench record & one row per rollout with its audited outcome label, from~\cite{cowling2026scaling} & primary set: standard (dim) lighting, placements N00 to N09 \\
Start frames & first frames of every recorded episode and of one bench session & frames 0 and 30; bench session of 29 September 2026 \\
Ball positions & start position of every ball, in millimetres relative to the front camera & per-frame grid fit; focal length 735 px; ball radius 14.4 mm (green), 15.6 mm (red) \\
\bottomrule
\end{tabularx}
\end{table}

Software: LeRobot commit 8352ca62 (v0.6.2). Statistics: Spearman's $\rho$ with permutation p-values (20,000 permutations; 10,000 within budgets), binomial simulation for the episode-count baseline (5,000 draws), Wilson intervals, Fisher's exact test.

\end{document}